Chapter

# Finding New Connections between Concepts from Medline Database Incorporating Domain Knowledge

*Yang Weikang, Chowdhury S.M. Mazharul Hoque and Jin Wei*

## Abstract

In this digital world, data is everything and significantly impacts our everyday lives. Interestingly, in this small world, everything is part of an ecosystem, where everything is connected, directly or indirectly. The same thing happens to data as well. In most cases, it may seem like a particular topic does not have any connection with another one, but in reality, they are connected through a mutually related topic. Therefore, in this research, we will discuss an adaptive model modified from the ABC model by Don R. Swanson, a Literature-Based Discovery (LBD) Model, to find the hidden connections between Concepts of Interest. The model demonstrates that two topics, "A" and "C," are different and have no relationship. But they have a common topic, "B," that can be used to connect topics "A" and "C." This famous model will be used in this discussion to connect Medical Concepts.

**Keywords:** medical data analysis, NLP, Medline database, text extraction, web mining

## 1. Introduction

In this era of technology, the use of data is increasing rapidly, and with the growth of AI, data is growing even faster, as it requires a lot of data to come to a conclusion about a given problem. Throughout the world, the volume of scientific literature is expanding immensely. Because of that, it has become very challenging to keep track of the recent innovations, even for the specialized researchers with their slender domains. From all those literature-based contents, extracting useful information has become a challenge to everyone, and this can be solved using Literature-Based Discovery. LBD minimizes the difficulty among the researchers by reducing the complexity regarding finding useful information.

In this paper, we have worked on an extended version of the existing ABC model, and we applied the model to a medical research article database, the MEDLINE database. The reason behind this research was that every day, the medical sector is growing, and there are new problems that are coming up. There are a lot of existing problems that need to be solved, and researchers are working on them. As a result, every year, a huge number of articles are generated, and for anyone, keeping track of the data or the most recent research or research outcome would be a challenge. In the year 1999, a simple logic-based model was proposed by Swanson, known as Complementary Structures in Disjoint Literature (CSD). In this model, he wanted to present that, in a research article, there can be many theories and findings that can be interconnected. For example, it is possible that in an article, it is mentioned that a certain medicine is used to treat the "Migraine Headache," and in another article, it says that that particular medicine commonly uses "Magnesium" as the main component [1]. Though "Migraine Headache" and "Magnesium" do not have a direct contact or relationship, and they exist in two different articles, it is still possible to make a relationship between those two.

This logic was explained by Swanson as follows – ""If concept A influences concept B, and concept B influences concept C, then concept A may influence concept C", and this model is also a Literature Based Discovery model called Swanson's ABC model [2]. Our research worked on an extended version of the ABC model. The model was applied to the Medline database, which is a database for worldwide medical-related research articles submitted to Medline and contains basic information, such as Title, Abstract, and Keywords. It is published by the National Library of Medicine (NLM). The proposed model uses the Pipe and Filter Software Architectural Design principles. The raw data was preprocessed to extract information that is relevant to the work before sending it to the final processing.

Moreover, the model was developed in such a way that it is not dependent only on two documents. Instead of that, it can go for cross-document discovery and produce a much better outcome. At the same time, the model was tested with a variety of cores to measure the impact on the result and processing time. It was found that the result on impact from the number of the cores was surprising. This model was tested and compared against other existing models and presented notable improvements. Data collected from the Medline database was processed using the MetaMap tool. It is a very popular tool that is used worldwide to process medical data, and it is also developed by NLM.

## 2. Swanson's ABC model

Don R. Swanson was the pioneer of Literature-Based Discovery, which is a form of knowledge extraction that can automatically generate hypotheses from a given dataset (basically a large amount of text data, such as abstracts of different scientific publications) to present new interdependence between existing concepts. The aim of LBD is to present indirect associations that are not totally direct. According to Don R. Swanson's hypothesis, two completely different studies that exhibit the A-B and B-C relationship may seem unchained, but an A-C relationship may emerge that was unexplored [3]. This process is also known as Swanson's Linking.

Don Swanson's original ABC model can be elaborated or explained through the "One Node Search" approach [4]. In this process, an initial topic or problem needs to be identified from an article that must have enough ground on the selected problem (Literature A). A second problem or topic needs to be selected from a second article with enough ground on the second topic and contains important information that would lead to the solution of problem A (Literature C). Here both articles have their own explanation and do not have any kind of interconnection. Now, from Literature A and C filtering important words and phrases needs to be carried out (B terms). B terms are the implicit information from both Literature A and C. Later, using those B terms ($B_1$, $B_2$, $B_3$,.........), searches need to be carried out to create $B_1$, $B_2$, $B_3$..... literature from Literature A and C. By scoring each $B_i$ literature from the title words and phrases from A literature it needs to figure out how many B literatures they are in. To define A literature, each $A_i$ term needs to be searched and analyzed. From the analysis, common $B_i$ terms from literature A and literature C would hypothetically contain common information related to the solution of the given problem.

Through this process, it has become much easier and faster to analyze and inspect indirect connections to gain new knowledge that has not been explicitly stated. This technique showed significant progress, and similar models were used to make new breakthroughs.

## 3. Literature review

The use of semantic predictions for LBD was presented and advocated by Hristovski, who was among the first few people who understand its importance [5]. To recover existing knowledge and find new relationships, he used the pattern-based approach that leveraged both term co-occurrence from the BioMedLEE and semantics from the SemRep [6–9]. To find possible connections, he used a specified priori-based discovery pattern. However, this model's major drawback was it could not be easily extended to find complex patterns. Therefore, interesting and more sophisticated patterns may remain untouched.

Delory Cameron et al. worked on Swanson's hypothesis using a graph-based recovery and decomposition model using semantic predictions [5]. As biomedical literature-based semantic predictions allow the development of labeled directed graphs and from the literature, various concepts can be associated, a methodology for the semi-automatic discovery model was developed. Their model was a new approach to the field and was able to find associations at multiple levels.

Another study was done by Erwan Moreau et al. on the literature-based discovery of the ABC model to find the limitations of the ABC model [10]. Their work presents that the ABC model can find a relationship among the contexts, but how far it is effective or not effective is unknown to everyone, and the ABC model does not have the ability to judge it from a technical point of view. Therefore, they have used a data-driven task to work as a middle ground between the quantitative and qualitative learning-based discoveries. However, the model has an issue with the relatively small predefined targets.

Sam Henry et al. worked on the biomedical domain with an extensive literature review of literature-based discovery, where a unifying framework was selected, and different models and methods were differentiated and elaborated [11]. The purpose of the model was to give ideas to the readers about different LBD publications, models, methodologies, and challenges. Finally, with that knowledge, users will be able to develop an idea to create their own model.

Another work on the multilevel context-specific relationship discovery model of biological data was developed by Sejoon Lee et al. Their research goal was to find the relationship between drugs and diseases [12]. Their model proposed a three-step procedure to find the expected outcome, and the steps were multi-level entity identification, interaction extraction from the literature, and context vector-based similarity score calculation. The outcome from their model presented that context-based relation inference can perform better than the traditional ABC model.

Researcher Yong Hwan Kim focused on the limitations of the ABC model as it only gives priority to the B entity only and on the way to the B entity it may grab some relationships that are not entirely valid [13]. To counter the limitations, they presented a context-based approach to connect entities.

## 4. Method

As this model uses Pipe and Filter Architecture Design, multiple layers of filters work together to produce the final outcome from the raw data. Each filter works independently by taking input data from the previous filter in a processable format and producing the desired output format for the next filter. The whole analysis can be divided into three parts- preprocessing, preparation phase, and output phase. The whole process maintains a data flow that is given in **Figure 1**. The tool used to process data is the MetaMap, which is a popular tool that has great implementation in the text extraction and web mining field [14, 15].

### 4.1 Data collection and preprocessing

As mentioned earlier, all the data was collected from the Medline database through an API. This database is maintained by the NLM, and researchers across the world can get free access to research articles. At this moment, more than 23 million articles are stored in the Medline database. However, the basic type of data available in the database is the title of the article and abstract. Data were downloaded in XML format and from those XML files based on specific tags, information such as Publication ID, Publication Date, Article Title, and Article Abstract was extracted and stored. A total of 746 Medline documents were extracted from the website, which is - ftp://ftp.ncbi.nlm.nih.gov/pubmed/baseline. Each of the Medline documents has 30,000 Medline references, and each reference points towards a research document. Therefore, in total, around 22,380,000 references were analyzed by the system. All the documents that were collected were in English, and if any of the titles were in another language, they were translated into English first. **Figure 2** shows a sample of the collected raw data.

### 4.2 Data processing through multiple filter

In this proposed model, filtering was done through multiple modules, and each module works independently. Here, the first module is the MedMeta module and it initially monitors the preprocessing. Later, it forwards the preprocessed data to the MetaMap API, where the MetaMap breaks down all the sentences in the document into phrases. In this case, those sentences are the title and abstract of the research articles. It normalizes those phrases to find mapping concepts through its database search in order to assign a mapping score to the concepts. Based on those concepts

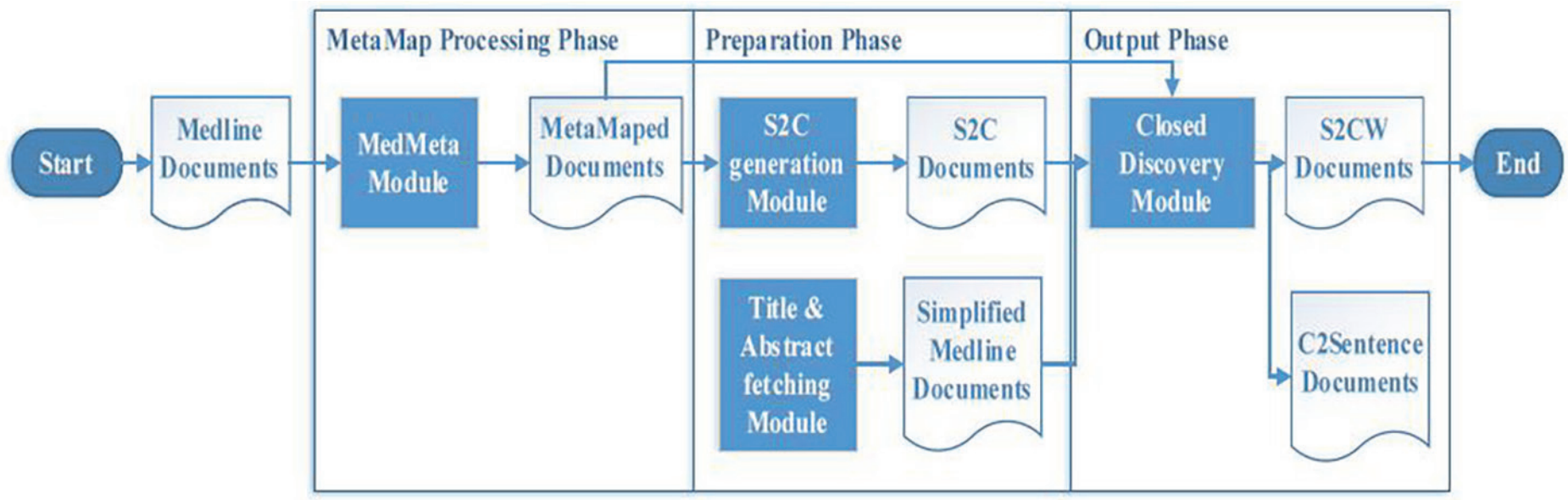

**Figure 1.**
*Data flow diagram.*

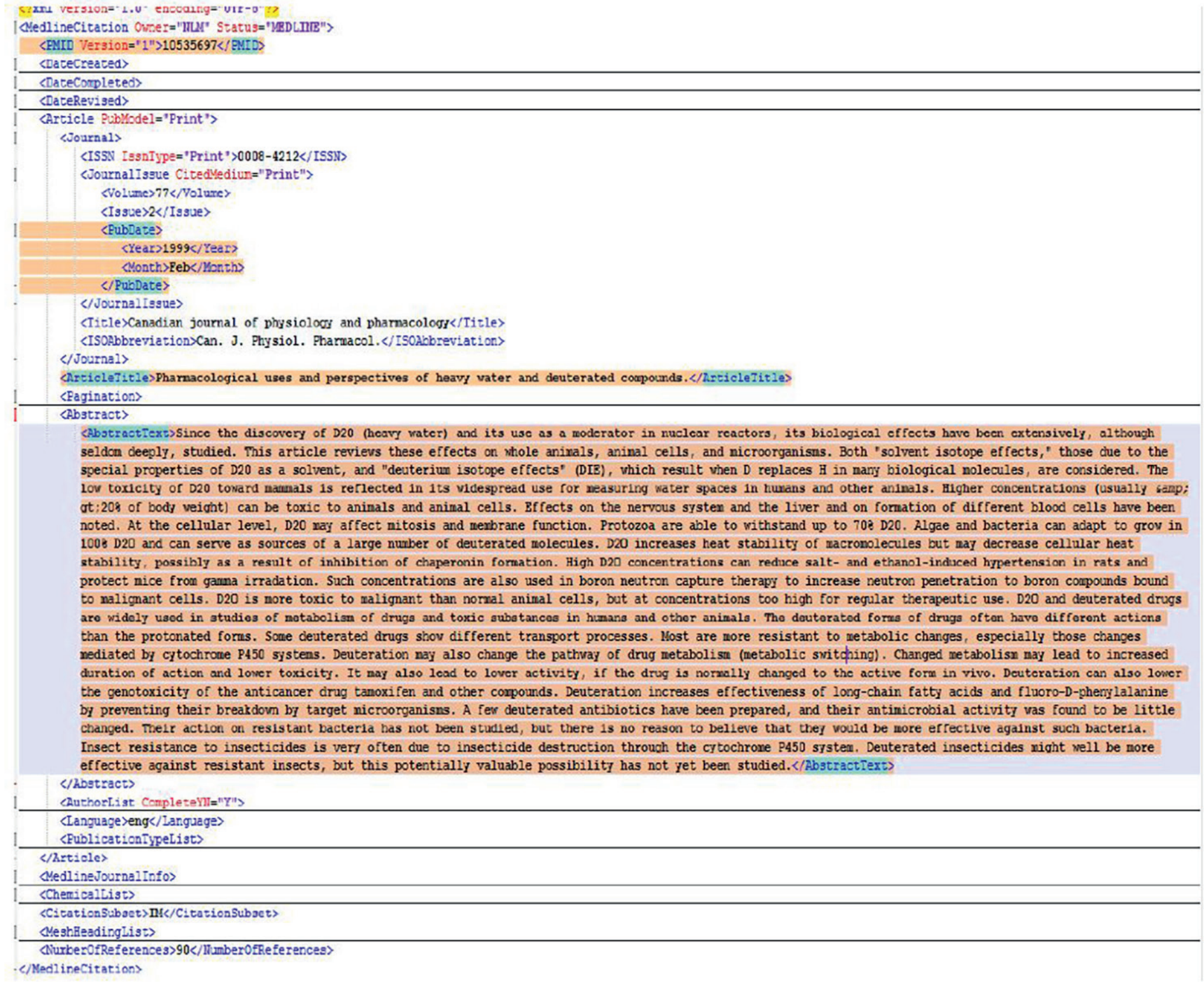

```xml
<?xml version="1.0" encoding="UTF-8"?>
<MedlineCitation Owner="NLM" Status="MEDLINE">
  <PMID Version="1">10535697</PMID>
  <DateCreated>
  <DateCompleted>
  <DateRevised>
  <Article PubModel="Print">
    <Journal>
      <ISSN IssnType="Print">0008-4212</ISSN>
      <JournalIssue CitedMedium="Print">
        <Volume>77</Volume>
        <Issue>2</Issue>
        <PubDate>
          <Year>1999</Year>
          <Month>Feb</Month>
        </PubDate>
      </JournalIssue>
      <Title>Canadian journal of physiology and pharmacology</Title>
      <ISOAbbreviation>Can. J. Physiol. Pharmacol.</ISOAbbreviation>
    </Journal>
    <ArticleTitle>Pharmacological uses and perspectives of heavy water and deuterated compounds.</ArticleTitle>
    <Pagination>
    <Abstract>
      <AbstractText>Since the discovery of D20 (heavy water) and its use as a moderator in nuclear reactors, its biological effects have been extensively, although
      seldom deeply, studied. This article reviews these effects on whole animals, animal cells, and microorganisms. Both "solvent isotope effects," those due to the
      special properties of D20 as a solvent, and "deuterium isotope effects" (DIE), which result when D replaces H in many biological molecules, are considered. The
      low toxicity of D20 toward mammals is reflected in its widespread use for measuring water spaces in humans and other animals. Higher concentrations (usually &amp;
      gt;20% of body weight) can be toxic to animals and animal cells. Effects on the nervous system and the liver and on formation of different blood cells have been
      noted. At the cellular level, D20 may affect mitosis and membrane function. Protozoa are able to withstand up to 70% D20. Algae and bacteria can adapt to grow in
      100% D20 and can serve as sources of a large number of deuterated molecules. D20 increases heat stability of macromolecules but may decrease cellular heat
      stability, possibly as a result of inhibition of chaperonin formation. High D20 concentrations can reduce salt- and ethanol-induced hypertension in rats and
      protect mice from gamma irradation. Such concentrations are also used in boron neutron capture therapy to increase neutron penetration to boron compounds bound
      to malignant cells. D20 is more toxic to malignant than normal animal cells, but at concentrations too high for regular therapeutic use. D20 and deuterated drugs
      are widely used in studies of metabolism of drugs and toxic substances in humans and other animals. The deuterated forms of drugs often have different actions
      than the protonated forms. Some deuterated drugs show different transport processes. Most are more resistant to metabolic changes, especially those changes
      mediated by cytochrome P450 systems. Deuteration may also change the pathway of drug metabolism (metabolic switching). Changed metabolism may lead to increased
      duration of action and lower toxicity. It may also lead to lower activity, if the drug is normally changed to the active form in vivo. Deuteration can also lower
      the genotoxicity of the anticancer drug tamoxifen and other compounds. Deuteration increases effectiveness of long-chain fatty acids and fluoro-D-phenylalanine
      by preventing their breakdown by target microorganisms. A few deuterated antibiotics have been prepared, and their antimicrobial activity was found to be little
      changed. Their action on resistant bacteria has not been studied, but there is no reason to believe that they would be more effective against such bacteria.
      Insect resistance to insecticides is very often due to insecticide destruction through the cytochrome P450 system. Deuterated insecticides might well be more
      effective against resistant insects, but this potentially valuable possibility has not yet been studied.</AbstractText>
    </Abstract>
    <AuthorList CompleteYN="Y">
    <Language>eng</Language>
    <PublicationTypeList>
  </Article>
  <MedlineJournalInfo>
  <ChemicalList>
  <CitationSubset>IM</CitationSubset>
  <MeshHeadingList>
  <NumberOfReferences>90</NumberOfReferences>
</MedlineCitation>
```

**Figure 2.**
*Sample XML file containing publication date, article ID, title and abstract.*

that were generated by the MedMeta server, indexes of concept occurrence relationships are produced and passed to the next filter.

The processes from the beginning to the end are monitored by one of the modules that can be called the MedMeta class. It observes the creation of the objects and connects them to the other objects. This model uses a parser to control event occurrence depending on a certain threshold or situation and the main reason behind it is that the amount of data to be processed can reduce the effectiveness of the model. For example, the meline documents were around 97 gigabytes in size, and an individual document reached up to 200 megabytes. The use of a Java-based purser reduced that pressure and ensured maximum effectiveness.

Another module named MedlineCite was responsible for storing data that was produced after parsing from the previous module. It divided the output data from the previous module into four categories, such as ID, Title, Abstract, and Date.

The next part of the module is the most important one, as it is responsible for communication with the MetaMap server in order to build indexes. A code sample for this part of the module is given below in **Figure 3**.

Titles and Abstracts will be passed through the server as input data to be processed, and the server will use Word Sense Disambiguation to separate the phrases and ignore any stop phrases. After processing the data, the MetaMap server will return the output. The generated output will be forwarded to the next filter to continue the analysis.

```
// Start a connection to submit the data, the default server is
// localhost
MetaMapApi api = new MetaMapApiImpl(host, port);
// Metamap behavior options
api.setOptions("-y -K -Q 0");

if (title != null) {
    //send title to MetaMap Server
    List<Result> tResults = api.processCitationsFromString(title);
    if (tResults.size() > 0) {
        tResult = tResults.get(0);
        p1 = new PostProcessor(tResult, PMID, pubDate, "T", semLst);
        p1.startPostProcess();
    }
}
if (articleAbstract != null) {
    //send abstract to MetaMap Server
    List<Result> aResults = api.processCitationsFromString(articleAbstract);
    if (aResults.size() > 0) {
        aResult = aResults.get(0);
        p2 = new PostProcessor(aResult, PMID, pubDate, "A", semLst);
        p2.startPostProcess();
    }
}

// disconnect the apiImplement to avoid open_too_many_files error
api.disconnect();
```

**Figure 3.**
*Code snippet for connecting MetaMap server and index creation.*

After that, the next filter will take the output from the previous filter as input and will gather important information. A three-level data structure (semantic class, concept class, and occurrence class) will hold the collected information. The task of the semantic class will be holding one or many concept objects. The concept class will hold one or many occurrence objects. Finally, the occurrence class is responsible for representing the occurrence of a certain word in a Medline document. It will hold information such as the genuine word that appears in the text, the original word preferable name, word location in the citation, Title, Abstract, ID of the citation and publication date where the word appears, and the exact location of the word that appaired on the title abstract. A sample of the MetaMapped document is given below in **Figure 4**.

Now, this three-level data structure produces XLM type of output document to hold those generated data that is shown in the picture.

At this end of the processing MedMeta module uses a multithread based parallel processing in order to improve the performance. As mentioned earlier, each Medline document holds 30,000 citations and those citations are divided into three parts, each having 10,000 citations. Each part will be processed by a single thread, therefore, total three threads will be working together at the same time to access those data. Each thread will be able to communicate with the MetaMap server individually and will produce separate results. When all three of the threads are done with the analysis, all three threads will merge, and all the results will be forwarded to the main thread to start producing the output. Due to the use of multiple threads, it was possible to present the impact of using multithread against a single thread in the evaluation.

Here, another module was used named Semantic Type to Concept (S2C Module), which is responsible for generating a simplified version of the MetaMapped document or S2C document. Previously a three-level data structure was used to process data, and now, frequently, two-level S2C relationships will be used. There is a reason behind it. If every time a request is made and against that every time a three-level relationship is transformed into a two-level relationship, it will take a lot of time to process the data.

```xml
<?xml version="1.0" encoding="UTF-8"?>
<Semantic name="aapp" description="Amino Acid, Peptide, or Protein">
    <Concept>
        <ConceptName>Oxidases</ConceptName>
        <TitleCount>1</TitleCount>
        <AbstractCount>0</AbstractCount>
        <Occurence>
            <OccurenceName>oxidases</OccurenceName>
            <PrefName>Oxidase</PrefName>
            <Tag>T</Tag>
            <PMID>16748552</PMID>
            <PubYear>1949</PubYear>
            <Offset>[(23, 8)]</Offset>
        </Occurence>
    </Concept>
    ...
</Sementic>
```

**Figure 4.**
*Sample MetaMapped document.*

Therefore, by using a two-level S2C relationship, document processing time can be reduced. The S2C generator module starts its job by looking for the file path to pass them through the module as input, and a parser builds a two-level relationship. The parser basically uses an event-driven technique to go through all the MetaMapped documents to find concepts and include them in semantic-type classes. The semantic type class maintains a Hash set and only unique values will be added to the hash set. In this research, there were 133 semantic types, and individually they had a Hash set of concepts. Outcomes were stored in an XML file to be processed by the next module. A sample S2C document is presented in **Figure 5**.

A special note should be mentioned that, as only the title and abstract are playing the main role here, it is not necessary to keep anything other than those two in the main processing part. However, other relevant data can be used to know more about the source of the concept. Therefore, the Medline document will be more simplified and ready for concept generation. A sample of the simplified Medline document is shown in **Figure 6**.

The final module in this model is called the ClosedDiscovery module, and it uses the MetaMapped Document, Simplified Medline Document, S2C Document, and the user-provided keyword as input. Those inputs will be used to generate Semantic Type to Concept and Weight (S2CW), which is the final output of the model, and it will be displayed through a Graphical User Interface (GUI).

The graphical user interface has two input boxes to provide input to the system, basically, those inputs are the keywords- Topic A and Topic C, whom the user wants to connect. To start the process, the user needs to click the start button on the screen (given in **Figure 7**). In the next stage, users will be able to select the S2C document

```xml
<?xml version="1.0" encoding="UTF-8" standalone="no"?>
<SemanticType name="aapp">
    <Concept>'Malic' enzyme</Concept>
    <Concept>(+)6a-hydroxymaackiain 3-O-methyltransferase</Concept>
    <Concept>(-)-Menthol dehydrogenase</Concept>
    <Concept>(123)I-mAb 14C5</Concept>
    <Concept>(125)I-L-PC</Concept>
    <Concept>(125)I-YLFQPQRFamide</Concept>
```

**Figure 5.**
*Sample S2C document.*

```
<MedLineCitation PMID="2447742">
    <Title>Latex agglutination test for alpha-fetoprotein in the diagnosis of premature
    rupture of the amniotic membranes (PROM).</Title>
    <Abstract>A rapid latex agglutination test for alpha-fetoprotein (AFP) was compared with
    a pH-indicator and patient history in the detection of premature rupture of the amniotic
    membranes. Of 120 patients examined, 34 had an established rupture of the membranes, and
    56 had suspected rupture. Thirty patients had no evidence of membrane rupture. The
    vaginal content was examined with a pH-indicator. Samples of vaginal content were also
    obtained to perform the latex agglutination test, and 103 of these samples were analysed
    by a radio-immunoassay (RIA) technique for AFP. Our results indicate that the latex
    agglutination test is of doubtful value due to the many inconclusive test results. The
    sensitivity of the latex test is less than 15%, and the specificity is 80%. Patient
    history combined with pH-indicator test is far more informative than the latex
    agglutination test.</Abstract>
</MedLineCitation>
```

**Figure 6.**
*A simplified Medline document.*

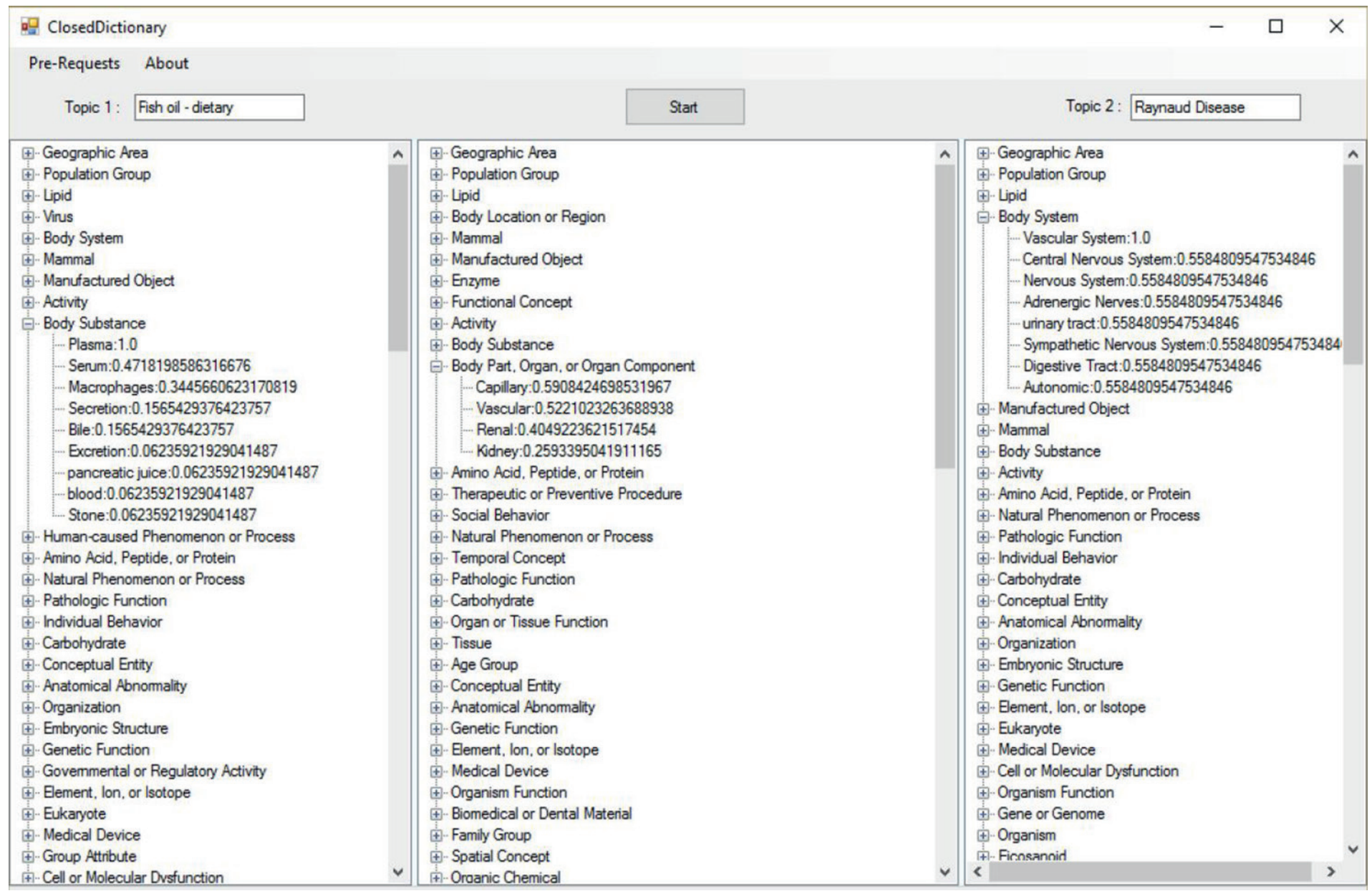

**Figure 7.**
*GUI for the model (showing input boxes for topics A and B, and their concepts, including the outcome C).*

that they want to be analyzed. Other S2C documents will not be considered for processing. As soon as the processing begins, it will start looking for the sentences relevant to the given keywords and will be stored in two different XML files based on their keywords (Topic A and C) (**Figures 8–10**).

A concept chain will be created from those data by calculating the Term Frequency (TF) and the Inverse Document Frequency (IDF) in the sentences. The appearance of a concept in all sentences is considered as its Term Frequency. Multiple appearances of a concept in the same sentence will be considered as one appearance.

The Inverse Document Frequency measures the concept's importance. For concept C it is measured by –.

$$IDF\left(c\right)=\log e\left(\frac{Total\ Number\ of\ Sentences}{Number\ of\ Sentences\ with\ concept\ C\ in\ it}\right) \tag{1}$$

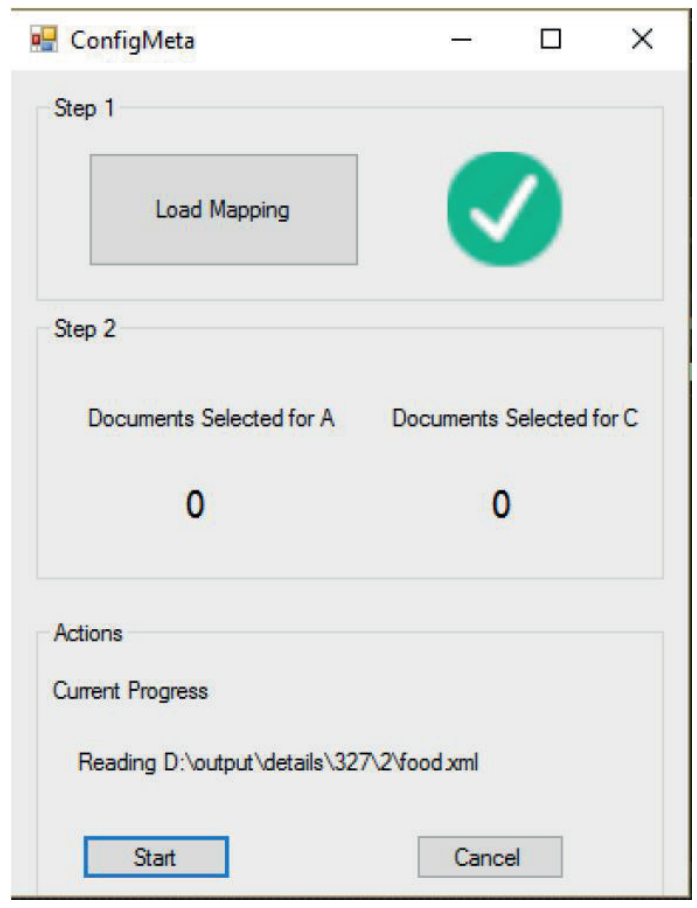

**Figure 8.**
*Page to selecting S2C documents for processing.*

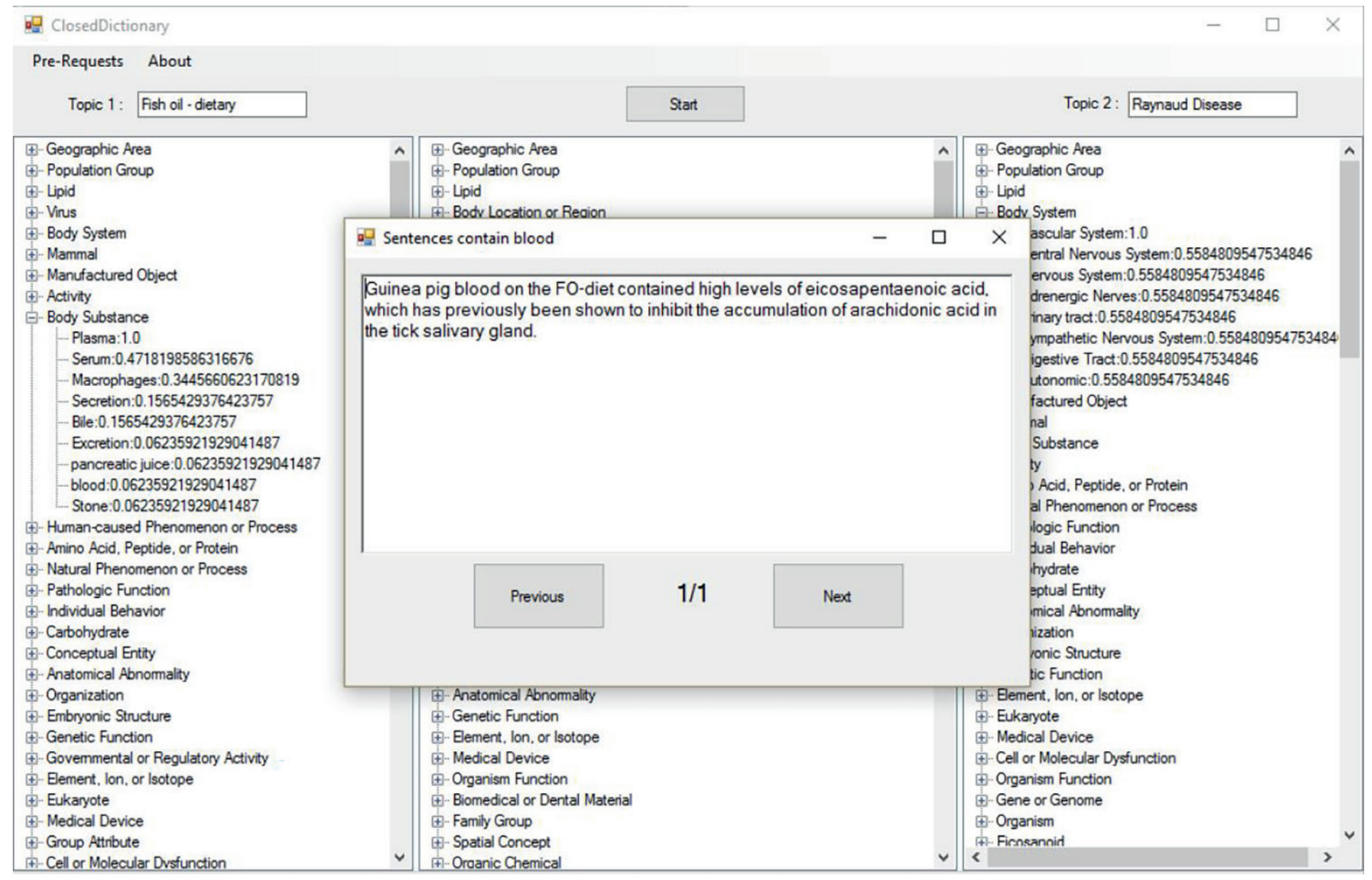

**Figure 9.**
*GUI showing all the relevant sentences.*

In theory the minimal appearance of a concept in a sentence leads to high IDF value, making it important to the context. Later, the concept weight will be calculated by multiplying the TF and IDF, which is shown in Eq. (2).

$$\text{Weight C} = TF\ \text{C}^{*}\ IDF\ \text{C} \tag{2}$$

As the results from the MetaMap server may contain multiple concepts, therefore, by iterating them it is possible to build the Concept to Weight (C2W) and Concept to Sentence (C2Sents) relationship. Now the S2C relationship will be joined with the C2W relationship to find the S2CW relationship, and the weight of each concept is normalized by

$$\text{Normalized Weight}(c) = \frac{\textit{the weight of the concept C}}{\textit{maximum weight in this Semantic Type}} \quad (3)$$

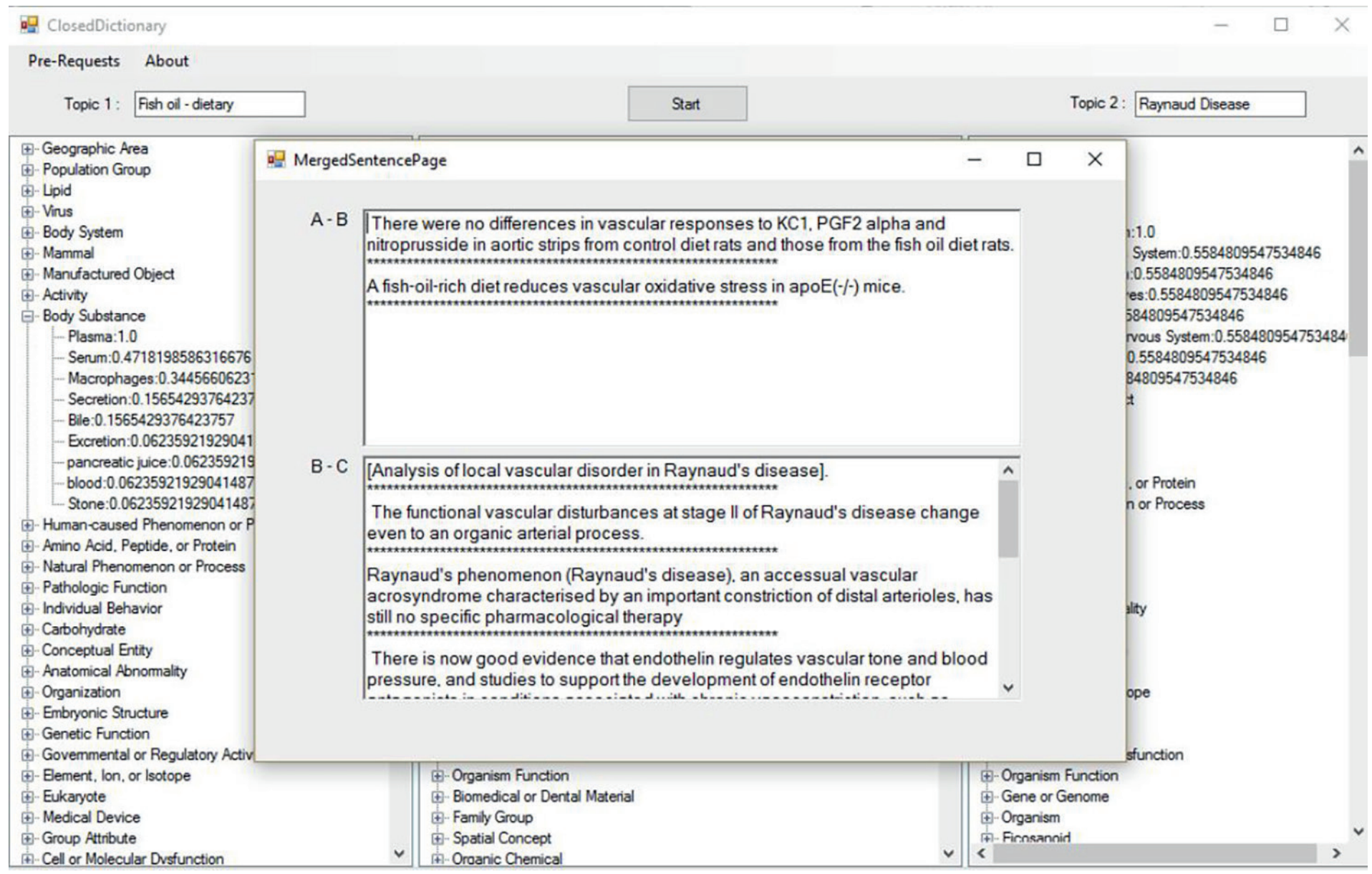

**Figure 10.**
*GUI presents sentences with intermediate linking concepts.*

Now to get the intermediate level information S2CW for topic B (S2CWB), S2CW for topic A (S2CWA) and S2CW for topic C (S2CWC) will be merged together. Potential connecting terms between topic A and B will be represented by S2CWB by crisscrossing their concepts. That means, all the concepts in S2CWB will be present in both S2CWA and S2CWC. The weight of S2CWA and S2CWB will determine the weight of S2CWB. The resulting S2CWB will be the relationship between topic A and topic C. The data was presented in the GUI, the left side represents the concepts and references for Topic A, and the right side represents the concepts and references for Topic C. In the center, topic B, or the relationship between input topics are presented.

### 4.3 System evaluation

There were quite a few modifications during the development of the MedMeta module. One of the major modifications was changing the model from a single-thread to a multi-thread model. Therefore, each of those threads can independently communicate with the MetaMap server, and the performance significantly improved and processing time reduced, though CPU use increased. An experiment was done to know how far performance changes due to changes in the number of threads used to communicate with the MetaMap server. The performance comparison is shown in **Table 1** below. However, in this research, a computer was used that has a Core i7 4th generation CPU with eight cores, and 16 GB of RAM with Windows 10 64-bit OS. The document used for the test had 180 Medline citations. Testing was performed on 1 thread, 2 threads and 3 threads.

| Number of threads | Processing time | Average memory usage | Average CPU usage | Performance compared to single thread | % of theoretical performance |
|---|---|---|---|---|---|
| 1 | 145 s | 98 MB | 25% | 1 | |
| 2 | 87 s | 110 MB | 40% | 1.67 | 85% |
| 3 | 62 s | 128 MB | 73% | 2.34 | 78% |

**Table 1.**
*Performance comparison on test data.*

According to **Table 1**, performance improved greatly based on the increment of the number of threads. First of all, processing time was reduced to 145 s (1 thread) to 87 s when 2 threads were used and again dropped to 62 s when 3 threads were used. On the other hand, performance increased 1.67 times while it was 85% of the theoretical performance when 2 threads were used, and 2.34 times while it was 78% of the theoretical performance when 3 threads were used. However, due to hardware limitations and as running multiple MetaMap servers costs a lot, the number of threads could not be increased, but with a computer with better configuration, it can be improved more.

### 4.4 Result analysis

From the given data set, the model found some meaningful connecting concepts based on the input topics A and C. All the relevant information was stored in the S2CWB XML files. But to find the accuracy of the result, the model was compared with another model by Gopalakrishnan and Kishlay [16]. They have used their own model on Mesh terms related to discovery from given topics, and connecting mesh terms were discovered that explain the relationship of input topics and found to be meaningful. The same topics, such as Fish-oil and Raynaud's Disease, were given to this Medline Module with the same dataset to test the accuracy.

#### *4.4.1 Test set 1 - fish oil and Raynaud disease*

In the first test set by Gopalakrishnan and Kishlay was on 'Fish oil and Raynaud Disease'. In their experiment, they found keywords such as Arthritis Rheumatoid, Platelet Aggregation, Prostaglandin, Blood Viscosity, and Vascular Reactivity (**Table 2**).

The table presents that out of five connecting words, four were found by the model. On the other hand, three of the connecting words gained higher rank in their semantic type. This model used article title and abstract to find the linking contexts, but model by Gopalakrishnan and Kishlay used Mesh terms as potential connecting terms, that were assigned by biomedical experts. However, this model also found some linking terms that were undetected by their model, such as "Hemodynamic" and "Atherosclerosis".

| Connecting concept | Find? | Weight | Rank in the semantic type |
|---|---|---|---|
| Platelet aggregation | True | 0.87 | 2 |
| Vascular reactivity | True | 1.27 | 1 |
| Blood viscosity | False, but found something about blood pressure | 0 | 0 |
| Prostaglandin | True | 0.309 | 6 |
| Arthritis rheumatoid | True, it appears as “Rheumatoid Arthritis” | 0.43 | 2 |

**Table 2.**
*Analysis of fish oil and Raynaud disease.*

### *4.4.2 Test set 2 - migraine and magnesium*

In the second study, two input topics were ‘Migraine and Magnesium’. In the experiment of Gopalakrishnan and Kishlay, they found connecting words like, Serotonin, Norepinephrine, Propranolol, Calcium, Ergotamine, Adenine Nucleotides, Adenosine Triphosphate, and Epinephrine (**Table 3**).

For Migraine and Magnesium this model was able to find all eight connecting words from the Gopalakrishnan and Kishlay’s model. Among those words, two of them got a very high semantic rank, and six of them were among the top ten semantic rank. Moreover, the model was able to find some connecting words, such as ‘Insulin’ that were not found previously.

### *4.4.3 Test set 3 - schizophrenia and phospholipase A2*

In Gopalakrishnan and Kishlay’s model ‘Schizophrenia and Phospholipase A2’ was used as the third test data and here the connecting words their model found are Trifluoperazine, Chlorpromazine, Prolactin, Choline, Norepinephrine, Arachidonic Acid, Receptors Dopamine and Phenothiazines (**Table 4**).

In this case study the result presented that all the connecting words from Gopalakrishnan and Kishlay’s model were found again. Among eight of those words, three gained very high semantic score. Moreover, again, this model was able to find connecting words like, “PGE2”, that were new to the model by Gopalakrishnan and Kishlay.

| Connecting concept | Find? | Weight | Rank in the semantic type |
|---|---|---|---|
| Propranolol | True | 0.41 | 8 |
| Adenosine triphosphate | True | 0.52 | 10 |
| Calcium | True | 1.50 | 1 |
| Ergotamine | True, the system found “Ergot Alkaloids” | 0.34 | 5 |
| Serotonin | True | 1.49 | 1 |
| Norepinephrine | True | 0.79 | 6 |
| Adenine nucleotides | True | 0.32 | 14 |
| Epinephrine | True | 0.49 | 13 |

**Table 3.**
*Analysis on migraine and magnesium.*

| Connecting concept | Find? | Weight | Rank in the semantic type |
|---|---|---|---|
| Chlorpromazine | True | 0.22 | 11 |
| Trifluoperazine | True | 0.07 | 32 |
| Receptors dopamine | True | 0.33 | 5 |
| Prolactin | True | 1.17 | 2 |
| Choline | True | 0.12 | 21 |
| Arachidonic acid | True | 1.18 | 1 |
| Phenothiazines | True | 0.05 | 56 |
| Norepinephrine | True | 0.3 | 11 |

**Table 4.**
*Analysis of schizophrenia and phospholipase A2.*

## 5. Future work

This model is currently discovering only one level of intermediate terms (chain length 1). In the future, our goal is to upgrade the model into multiple levels, so that when there is only a small amount of information available, it becomes easier to find possible connections. Moreover, MetaMapped files will be combined together for easier access to the files. Finally, other MetaMap options will be explored to find the one that can provide the best result.

## 6. Conclusion

This research discussed the Learning Based Discovery (LBD) models and, more specifically, one of the LBD models, which is the ABC model. An extended version of the ABC model was presented with a complete application. This era of technology is growing very fast, and with its growth, the use of data is also increasing. Therefore, sometimes, it is difficult to find relevant data about a particular topic. This complexity can be reduced by the use of LBD models. There is no LBD model that can be considered as the best model, because all the systems have their problems and benefits. A model may not provide the best result for a particular problem, but can perform amazingly for another. Moreover, interaction with the user interface makes any system easier. Because of that, the popularity of UI-based systems is increasing among researchers as well. However, this modified Swanson's ABC model can be improved more to solve problems faster and with a lot more data. This type of model will be able to make things easier for researchers by finding relevant data based on its user's input data from a larger data set.

## Acknowledgements

This research is supported by the National Science Foundation award IIS-1739095.

## Author details

Yang Weikang[1], Chowdhury S.M. Mazharul Hoque[2]* and Jin Wei[2]

1 North Dakota State University, Fargo, USA

2 University of North Texas, Denton, Texas, USA

*Address all correspondence to: smmazharulhoquechowdhury@my.unt.edu